\documentclass[10pt,twocolumn,letterpaper]{article}
\pdfoutput=1
\usepackage{cvpr}              

\usepackage[dvipsnames]{xcolor}

\usepackage{url}
\usepackage{times}
\usepackage{epsfig}
\usepackage{graphicx}
\usepackage{amsmath}
\usepackage{amssymb}
\usepackage{multirow}
\usepackage{cases}
\usepackage{booktabs}
\usepackage{color}
\usepackage{cases}
\usepackage{wasysym}

\definecolor{cvprblue}{rgb}{0.21,0.49,0.74}
\usepackage[pagebackref,breaklinks,colorlinks,citecolor=cvprblue]{hyperref}
\def\paperID{6895} 
\def\confName{CVPR}
\def\confYear{2024}

\title{Scale Decoupled Distillation}

\author{Shicai Wei \quad\quad\quad  Chunbo Luo \quad\quad\quad  Yang Luo  \\
School of Information and Communication Engineering \\
University of Electronic Science and Technology of China\\
{\tt\small shicaiwei@std.uestc.edu.cn  \{c.luo, luoyang\}@uestc.edu.cn }}

\begin{document}
\maketitle

\begin{abstract}

Logit knowledge distillation attracts increasing attention due to its practicality in recent studies. However, it often suffers inferior performance compared to the feature knowledge distillation. In this paper, we argue that existing logit-based methods may be sub-optimal since they only leverage the global logit output that couples multiple semantic knowledge. This may transfer ambiguous knowledge to the student and mislead its learning. To this end, we propose a simple but effective method, i.e., Scale Decoupled Distillation (SDD), for logit knowledge distillation. SDD decouples the global logit output into multiple local logit outputs and establishes distillation pipelines for them. This helps the student to mine and inherit fine-grained and unambiguous logit knowledge. Moreover, the decoupled knowledge can be further divided into consistent and complementary logit knowledge that transfers the semantic information and sample ambiguity, respectively. By increasing the weight of complementary parts, SDD can guide the student to focus more on ambiguous samples, improving its discrimination ability. Extensive experiments on several benchmark datasets demonstrate the effectiveness of SDD for wide teacher-student pairs, especially in the fine-grained classification task.  Code is available at: \href{https://github.com/shicaiwei123/SDD-CVPR2024}{https://github.com/shicaiwei123/SDD-CVPR2024}

\end{abstract}

\section{Introduction}

Knowledge distillation is a general technique for assisting the training of “student” networks via the knowledge of pre-trained “teacher” networks~\cite{kds}. Depending on the location of transferred knowledge, distillation methods are divided into two categories: logit-based distillation~\cite{KD1} and feature-based distillation~\cite{hint}.  Due to the computational efficiency~\cite{dkd} and ability to handle heterogeneous knowledge~\cite{wei2023privileged}, logit distillation has gained increasing attention in recent years. 

\begin{figure}[t]
\centering
\includegraphics[width=1.0\columnwidth]{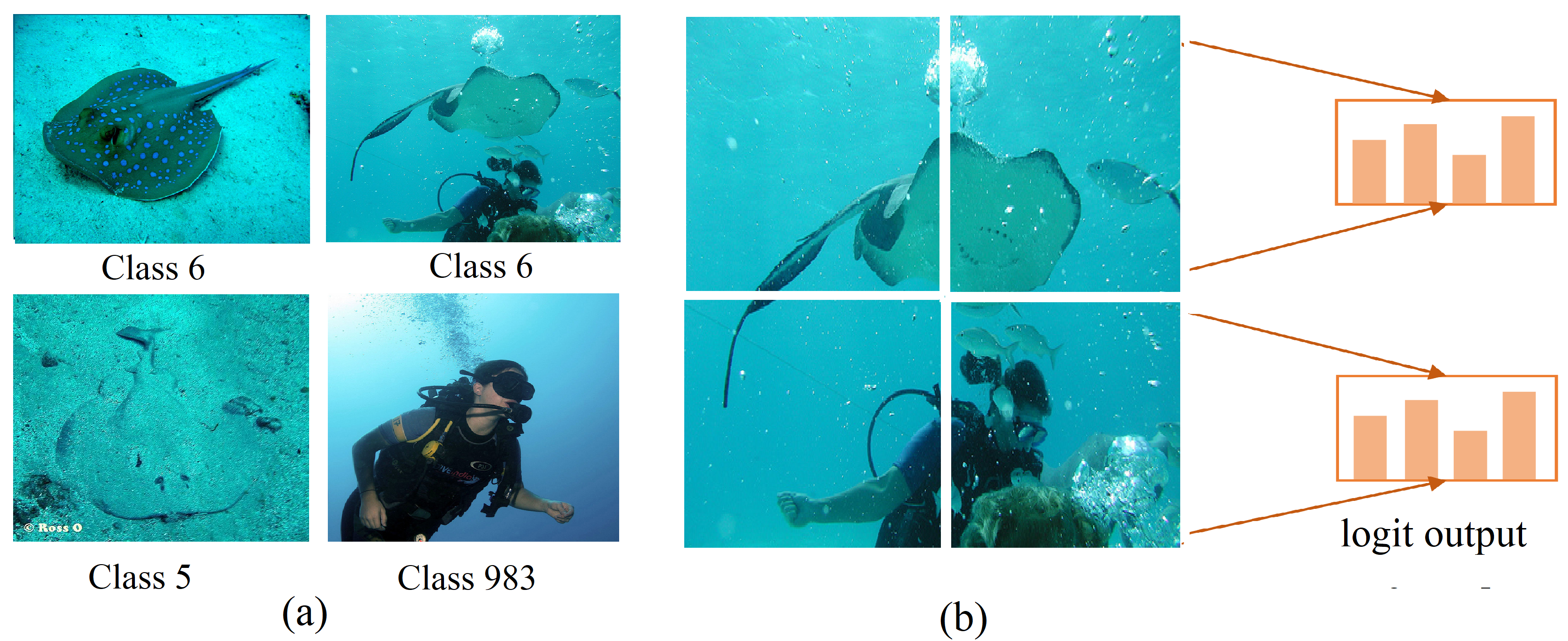} 
\caption{Image visualization on ImageNet. (a) The top line shows some misclassified samples of class 6 in ResNet34. The bottom line displays their corresponding predicted class and sample. (b) Illustrates the intuitive model for scale decoupling.}
\label{case_prediction}
\end{figure}

To date, many logit distillation methods have been proposed and can be broadly categorized into two groups. The first group aims to extract richer logit knowledge by introducing multiple classifiers~\cite{dcm,one} or through self-supervision learning~\cite{sskd}. The second group aims to optimize the knowledge transfer by techniques like dynamic temperature~\cite{fn} or knowledge decoupling~\cite{dkd,nkd, KDExplainer}. While these methods achieve good results, we argue they could lead to sub-optimal results as they solely rely on the global logit knowledge of the whole input.




Specifically, the whole image usually couples the information of multiple classes and leads to misclassification. On one hand, two classes may belong to the same superclass, sharing similar global information in their samples. Illustrated in the first column of Fig.~\ref{case_prediction}(a), both classes 5 and 6 belong to the superclass "fish" and exhibit similar shapes. Additionally, as shown in the second column of Fig.~\ref{case_prediction}(a), a scene may encompass information from multiple classes, such as classes 6 and 983, creating a logit output that is semantically mixed. Consequently, the global logit output fuses diverse and fine-grained semantic knowledge. This may transfer the ambiguous knowledge to the student and mislead its learning, resulting in sub-optimal performance.

To this end, we propose the SDD method to assist the logit distillation by decoupling the logit output at the scale level. Specifically, as shown in Fig.~\ref{case_prediction}(b), SDD decouples the logit output of the whole input into the logit outputs of multiple local regions. This helps acquire richer and unambiguous semantics knowledge. Then SDD further divides the decoupled logit outputs into consistent and complementary terms according to their class. Consistent terms belong to the same class as the global logit output, transferring multi-scale knowledge of the corresponding class to students. Complementary terms belong to classes different from the global logit output, preserving sample ambiguity for the student and avoiding overfitting to ambiguous samples. Finally, SDD performs the distillation among all these logit outputs to transfer comprehensive knowledge from the teacher to the student, improving its discrimination ability to ambiguous samples.

In total, we summarize our contributions and the differences from the existing approaches as follows:

\begin{itemize}
  \item We reveal a limitation of classic logit distillation resulting from the coupling of multi-class knowledge. This hinders students from inheriting accurate semantic information for ambiguous samples.
  \item We propose a simple but effective method, i.e., SDD, for logit knowledge distillation. SDD decouples the global logit output as consistent and complementary local logit output and establishes the distillation pipeline for them to mine and transfer richer and unambiguous semantic knowledge.
  \item We present extensive experiments on several benchmark datasets, demonstrating the effectiveness of SDD for a wide range of teacher-student pairs, particularly in fine-grained classification tasks.

\end{itemize}


\section{Related Work}

\subsection{Feature-based Distillation}. The feature-based distillation is first proposed in Fitnets~\cite{hint}, in which the student is trained to mimic the output of the intermediate feature map of the teacher directly. Since then, a variety of other methods have been proposed to extend the fitnets by matching the features indirectly. For example, AT~\cite{hint1} distills the attention map of the sample feature from a teacher to a student. Besides, some methods are further proposed to transfer the inter-sample relation, such as RKD~\cite{relation-rkd}, SP~\cite{similar}, and reviewKD~\cite{review}. While these methods achieve good results, their performance usually degrades when handling teacher-student pairs with heterogeneous architecture, especially those with different layers~\cite{information-flow,SemCKD}. Therefore, this paper pays attention to the logit-based distillation that has good generalization for heterogeneous knowledge distillation

\subsection{Logit-based Distillation}. The logit-based distillation is originally proposed by Hinton~\cite{KD1}, in which the student is trained to mimic the soft logit output of the teacher. Then several methods are proposed to promote its performance. FN~\cite{fn} introduces the L2-norm of the feature as the sample-specific correction factor to replace the unified temperature of KD. SSKD~\cite{sskd} trains extra classifiers via the self-supervision task to extract “richer dark knowledge” from the pre-trained teacher model. KDExplainer~\cite{KDExplainer} proposes a virtual attention module to improve the logit distillation by coordinating the knowledge conflicts for discriminating different categories. WSLD~\cite{wlsd} analyzes soft labels and distributes different weights for them from a perspective of bias-variance trade-off. SSRL~\cite{ssrl} transfers the knowledge by guiding the teacher’s and student’s features to produce the same output when they pass through the teacher’s pre-trained and frozen classifier. SimKD~\cite{simKD} transfers the knowledge by reusing the teacher's classifier for the student network. In addition, the latest DKD~\cite{dkd} proposes the decoupled knowledge distillation that divides the logit knowledge into target knowledge and non-target knowledge. NKD~\cite{nkd} further proposes to normalize the non-target logits to equalize their sum. However, all of them only focus on the global logit knowledge with mixed semantics, transferring the ambiguous knowledge to the student.

\section{Method}
\label{SDD} 

In this section, we revisit the conventional KD and then describe the details of the proposed scale-decoupled knowledge distillation.


\textbf{Notation.} Given an image input $x$, let $T$ and $S$ denote the teacher and student networks, respectively. We split these networks into two parts: (i) one is the convolutional feature extractor $f_{Net}, Net = \left\{T, S\right\}$, then the feature maps in the penultimate layer are denoted as $f_{Net}(x) \in R^{c_{Net} \times h_{Net} \times w_{Net}}$, where $c_{Net}$ is the number of feature channels, $h_{Net}$ and $w_{Net}$ are spatial dimensions. (ii) Another is the projection matrix $W_{Net} \in R^{c_{Net} \times K}$, which project the feature vector extracted from $f_{Net}(x)$ into $K$ class logits $z_{Net}^{l}$, $l=1,2,...,K$. Then, let $f_{Net}(j,k)= f_{Net}(x) (:,j,k) \in R^{ c_{Net} \times 1 \times 1}$ denotes the feature vector at the location $(j,k)$ of the $f_{Net}(x)$. According to the receptive field theory in~\cite{spp-net,fast-rcnn}, $f_{Net}(j,k)$ can be regarded as the representation of the region $(t_{x},t_{y},t_{x}+d,t_{y}+d)$ in $x$, where $t_{x}=d*j$, $t_{y}=d*k$ and $d$ is the downsampling factor between the input and the final feature map.

\begin{figure*}[t]
\centering
\includegraphics[width=1.0\textwidth]{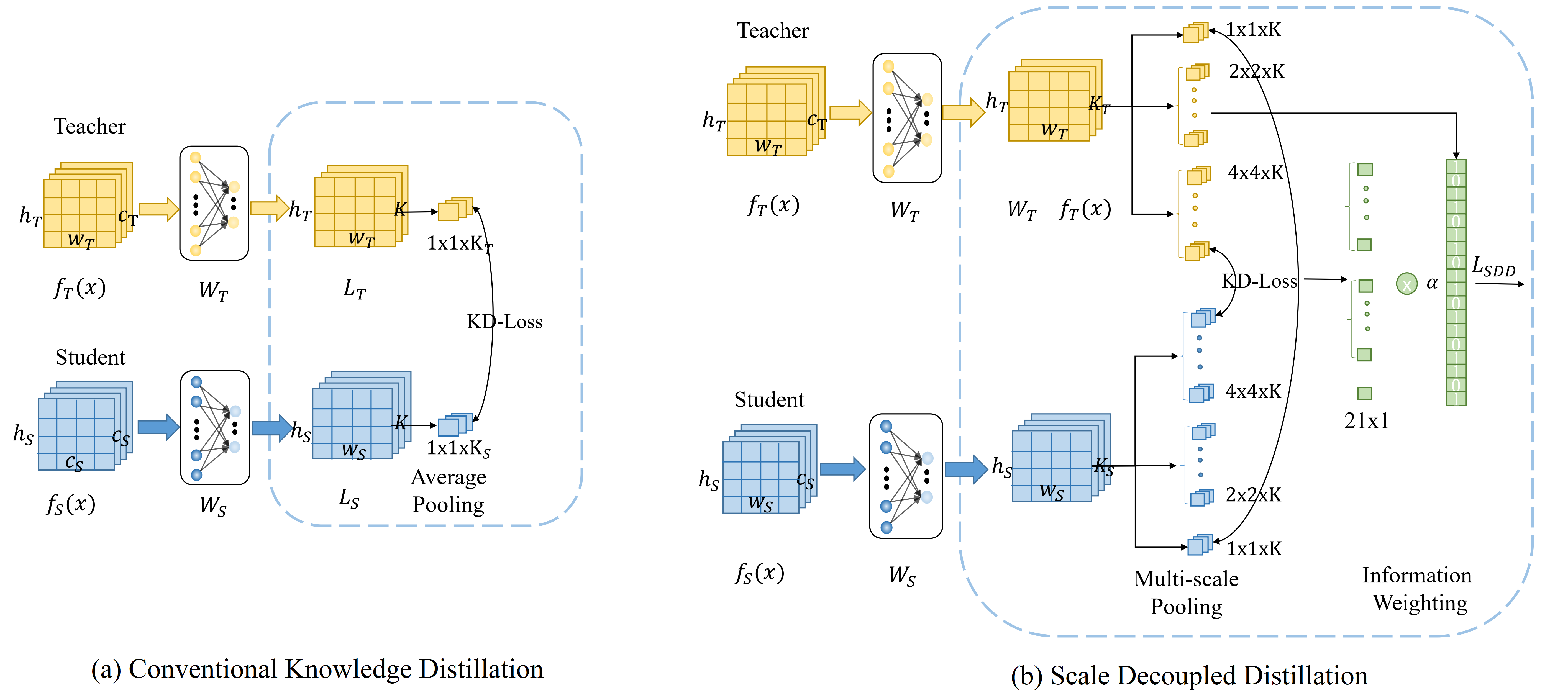} 
\caption{Illustration of the conventional KD (a) and our \textbf{SDD (b)}. Compared with the conventional KD that only considers the global logit knowledge via global average pooling, SDD proposes to capture the multi-scale logit knowledge via the multi-scale pooling so that the student can inherit the fine-grained and unambiguous semantic knowledge from the teacher. }
\label{scale}
\end{figure*}

\subsection{Conventional Knowledge Distillation}

The concept of knowledge distillation is first proposed in~\cite{KD1} to distill the logit knowledge from the teacher to the student by the following loss,


\begin{numcases} 
 \mathcal{L}_\mathcal{KD}=\mathcal{KL}(\operatorname{\sigma}(P_{T}) \| \operatorname{\sigma}(P_{S}))\\ 
 P_{T}=W_{T}\sum\limits_{j=0}^{h_{T}-1}\sum\limits_{k=0}^{w_{T}-1}\frac{1}{h_{T}w_{T}}f_{T}(j,k)\\ 
 \label{E3}
  P_{S}=W_{S}\sum\limits_{j=0}^{h_{S}-1}\sum\limits_{k=0}^{w_{S}-1}\frac{1}{h_{S}w_{S}}f_{S}(j,k) 
\end{numcases} where $\sigma(.)$ is the softmax function and $\mathcal{KL}(.,.)$ means the KL divergence. 

Due to the linearity of fully connected layer, $P_{T} $ and $P_{S}$ can be rewrite as follows,
\begin{numcases}
    \mathcal{P}_{T}= \sum\limits_{j=0}^{h_{T}-1}\sum\limits_{k=0}^{w_{T}-1}\frac{1}{h_{T}w_{T}} {L_{T}}(j,k)\\ 
    \label{Lt}
  P_{S}= \sum\limits_{j=0}^{h_{S}-1}\sum\limits_{k=0}^{w_{S}-1}\frac{1}{h_{S}w_{S}}{L_{S}}(j,k)
  \label{Ls}
\end{numcases} where $L_{T}=W_{T} f_{T}(x)$ and $L_{S}=W_{S} f_{S}(x)$ mean the logit output maps of teacher and student, respectively.

From the above equations, we can see that the conventional logit-based distillation only leverages the average logit output that mixes different local logit knowledge calculated from different local feature vectors, such as ${L_{T}}_{11}$. However, as shown in Fig.~\ref{case_prediction}(a), different local outputs usually contain distinct semantic information. Simply fusing them in a logit output would transfer ambiguous knowledge to the student and mislead its learning. Consequently, conventional logit-based distillation usually leads to sub-optimal performance.

To overcome this limitation, we propose the SDD that decouples the logit output at the scale level to mine richer and unambiguous logit knowledge for student learning.



\subsection{Scale Decoupled Knowledge Distillation}
\label{aaaaaaaaaaaaa}

In this section, we describe the proposed SDD in detail. As shown in Fig.~\ref{scale}(b), the SDD consists of two parts: multi-scale pooling, and information weighting. Specifically, given the logit output maps of teacher and student, i.e. $L_{T}$ and $L_{S}$, multi-scale pooling runs the average pooling on different scales to acquire the logit output of different regions of the input image. Compared with the conventional KD that only considers global logit knowledge, this helps preserve fine-grained knowledge with clear semantics for the student. Then, we establish the knowledge distillation pipeline for the logit output at each scale. Finally, information weighting increases the weight of the distillation loss for the local logit that has inconsistent classes with the global logit. This guides the student network to pay more attention to the ambiguous samples whose local and global categories are inconsistent.



Specifically, the multi-scale pooling splits the logit output map into cells at different scales and performs average pooling operations to aggregate the logit knowledge in each cell. Let $\mathcal{C}(m,n)$ denotes the spatial bins of the $n_{th}$ cell at $m_{th}$ scale, $\mathcal{Z}(m,n)$ denotes the input region corresponding to this cell, ${\pi}_{T}(m,n) \in R^{ K\times 1 \times 1 }$ denotes the logit output of the teacher for region $\mathcal{Z}(m,n)$, which is the aggregated logit knowledge of this cell,

%

\begin{equation}
    \mathcal{\pi}_{T}(m,n)=\sum\limits_{j,k \in \mathcal{C}(m,n)}\frac{1}{m^{2}}L_{T}(j,k)
\end{equation} 
where $(j,k)$ means the coordinate of the logit output in $\mathcal{C}(m,n)$. And the paired logit output of the student for the same region $\mathcal{Z}(m,n)$ is the ${\pi}_{S}(m,n) \in R^{K \times 1 \times 1 }$,






\begin{equation} 
\pi_{S}(m,n)=\sum\limits_{j,k \in \mathcal{C}(m,n)}\frac{1}{m^{2}}L_{S}(j,k)
\end{equation} 
where $m$ and $n$ are as same as those in ${\pi}_{T}(m,n)$. For each paired logit output, the distillation loss $\mathcal{D}(m,n)$ that transfers the logit knowledge at the region $\mathcal{Z}(m,n)$ from the teacher to the student is defined as follows,

\begin{equation}
\label{lr}
\begin{split}
\mathcal{D}_(m,n) &=\mathcal{LD}(\operatorname{\sigma}({\pi}_{T}(m,n)), \operatorname{\sigma}({\pi}_{S}(m,n)))
\end{split}
\end{equation}
where $\mathcal{LD}(.,.)$ denotes the conventional logit-based distillation loss, such as the KL divergence in~\cite{KD1} and the decoupling loss in~\cite{dkd}. By traversing all the scales $m$ in ${M}=\left\{1,2,4,...,w \right\}$ and their corresponding cells $N_{m}=\left\{1,4,16,...,w^{2}\right\}$, we can get the final SDD loss as follows,



\begin{equation}
\begin{split}
\mathcal{L}_{SDD}  &=\sum_{m \in M}\sum_{n \in N_{m}}\mathcal{D}(m,n)
\end{split}
\end{equation}



Besides, we can further divide the decoupled logit outputs into two groups via their classes. One is the consistent terms that belong to the same class with the global logit output. Another is the complementary terms that belong to different classes from the global logit output. Here, the consistent terms transfer the multi-scale knowledge of corresponding classes to students. The complementary terms preserve the sample ambiguity for the student. Specifically,  when the global prediction is right while the local prediction is wrong, the inconsistent local knowledge encourages the student to preserve the sample ambiguity, avoiding overfitting the ambiguous samples. On the other hand, when the global prediction is wrong while the local prediction is right, inconsistent local knowledge can encourage the student to learn from similar components among different categories, alleviating the bias caused by the teacher.

Here, we introduce independent hyper-parameters for complementary terms to control the level of regularization and ${L}_{SDD}$ can be rewritten as follows,

\begin{equation}
\begin{split}
\mathcal{L}_{SDD}  &= \mathcal{D}_{con} + \beta \mathcal{D}_{com}
\end{split}
\end{equation} where $ \mathcal{D}_{con}$ and $\mathcal{D}_{com}$ denotes the loss sum of consistent and complementary logit knowledge respectively.



Finally, combined with label supervision, the total training loss for the student to leverage multi-scale patterns from the teacher to improve its performance is defined as follows,
 \begin{equation}
 \label{l1}
 \mathcal{L}_{1}= \mathcal{L}_{CE}+\alpha \mathcal{L}_{SDD} 
\end{equation}
where $\mathcal{L}_{CE}(.,.)$ denotes the label supervision loss for the task at hand, and $\alpha$ is a balancing factor. 

\textbf{Compared with conventional KD}: Particularly, $\mathcal{L_SDD}$  will degenerate into conventional distillation loss considering global logit knowledge when  $m=w_{T}=h_{T}$, $n=1$. Thus, the conventional knowledge distillation loss can be regarded as a term of the SDD loss, covering the entire image (0,0,$w*d$,$w*d$). It encourages the student to learn the contextual information of the whole image from the global logit output of the teacher. Besides, when $m<w_{T}=h_{T}$, SDD further calculates the logit output at local scale loss to preserve the fine-grained semantic knowledge. These terms guide the student to inherit diverse and clear semantics knowledge from the teacher, enhancing its discrimination ability to the ambiguous samples. 

\textbf{Compared with multi-branch KD}: Multi-branch KD, such as ONE~\cite{one} and DCM~\cite{dcm}, usually consists of multiple classifiers and fuses their logit output as the teacher knowledge to guide the training of the student. This introduces additional structure complexity. In contrast, although SDD introduces multi-scale pooling to generate multi-scale features, it calculates the multi-scale logit output via the same classifier. In other words, SDD is still the single-branch method and does not introduce any extra structural and computational complexity.


\textbf{Compared with DKD}: DKD~\cite{dkd} decouples the logit knowledge as the target class and non-target class knowledge, which is conducted on the class scale and performed after calculating the logit output. In contrast, SDD decouples the logit knowledge as multi-scale knowledge, which is conducted on the spatial scale and performed before calculating the logit output. In particular, DKD can also be embedded in SDD and achieves better performance (see Table~\ref{cifar100_diff}, Table~\ref{imagenet-diff}).

\begin{table*}[ht]
\centering
\begin{tabular}{ccccc}
\toprule
Teacher  & ResNet32x4  & WRN40-2               & WRN40-2     & ResNet50     \\ 
         & 79.42       & 75.61                 & 75.61       & 79.34        \\ \toprule
Student  & MobileNetV2 & VGG8                  & MobileNetV2 & ShuffleNetV1 \\ 
         & 64.6        & 70.36                 & 64.6        & 70.50        \\ \toprule
FitNet   & 65.61       & 70.98                 & 65.12           & 72.03        \\
SP       & 67.52       & 73.18                 & 66.34           & 71.28        \\ 
CRD      & 69.13       & 73.88                 &68.89         & 75.70        \\ 
SemCKD   & 68.99       & 73.67                 & 68.34          & 75.56        \\ 
ReviewKD & -           & -                     & -           & -            \\ 
MGD      & 68.13       & 73.33                 & 68.55         & 74.99        \\ \toprule
KD       & 67.72       & 73.97                 & 68.87       & 75.82        \\ 
SD-KD    & 68.84(+\textbf{1.12})       &74.44(+ \textbf{0.57}) & 69.81(+\textbf{0.94})        & 76.87(+\textbf{1.05}) \\ 
DKD      & 68.98       & 74.13                 & 69.33       & 77.01        \\ 
SD-DKD   & \textbf{70.08(+\textbf{1.1})}       & \textbf{74.58(+\textbf{0.45})  }        &\textbf{ 70.13(+\textbf{0.8})}       & \textbf{78.11(+\textbf{1.1})}  \\ 
NKD      & 68.81       & 73.80                & 68.85       &     76.22         \\ 
SD-NKD   & 69.59(+\textbf{0.78})       & 74.34(+\textbf{0.54})           & 70.04(+\textbf{1.19})       &    76.95(+\textbf{0.73})        
\\ \toprule
\end{tabular}
\caption{Performance of model compression on the CIFAR-100 dataset. Here, the teacher and student with different network structures and layers. Specifically, the layers of ResNet32x4, WRN40\_2, and ResNet50 are 4, 4, and 3 while the layers of MobileNetV2, VGG8, and ShuffleNetV1 are 3, 3, and 4, respectively.}
\label{diff_all}
\end{table*}

\section{Experiments}



\subsection{Experimental Setups}

\textbf{Datasets.} We conduct the experiments on the CIFAR100~\cite{cifar100}, CUB200~\cite{cub200}, and ImageNet~\cite{imagenet}. Here, CIFAR100 and ImageNet are used for the evaluation of classic classification tasks. CUB200 is used for the evaluation of fine-grained classification tasks, which includes 200 different species of birds.

\textbf{Implementation Details.} As described in Section~\ref{aaaaaaaaaaaaa}, the $\mathcal{LD}$ used for $L_{SDD} $ can be implemented with arbitrary logit-based loss. Here, for a fair comparison with previous logit-based methods and to study the effectiveness of the SDD, we use the same losses used in the KD, DKD, and NKD methods and denote these implementations as \textbf{SD-KD} \textbf{SD-DKD}, and \textbf{SD-NKD}, respectively.

SDD has three hyper-parameters, scale set $M$, balance parameters $\alpha$ and $\beta$ The detailed analysis of $M$, and $\beta$ can be seen in the ablation study. Specifically, ${M}$ is set as $\left\{1,2,4\right\}$ for the distillation tasks with different architectures, and $\left\{1,2\right\}$ for the distillation tasks with similar architectures. $\beta$ is set as 2.0. As for the $\alpha$, for a fair comparison with previous logit-based methods, we follow the same setting used in the original KD, DKD, and NKD methods. In particular, since the sum of multi-scale logit distillation loss could be high and lead to a large initial loss, we utilize a 30-epoch linear warmup for all experiments.

\begin{table*}[h]

\begin{center}
\centering

\begin{tabular}{ccccccc}
\toprule
Teacher  & ResNet32x4   & WRN40-2      & ResNet50     & VGG13        & ResNet32x4   & ResNet50      \\ 
Acc      & 79.42        & 75.61        & 79.34        & 74.64        & 79.42        & 79.34         \\ \toprule
Student  & ShufleNetV1  & ShufleNetV1  & MobileNetV2  & MobileNetV2  & ShufleNetV2  & VGG8          \\ 
Acc      & 70.50        & 70.50        & 64.6         & 64.6         & 71.82        & 70.36         \\ \toprule
FitNet   & 73.54        & 73.73        & 63.16        & 63.16        & 73.54        & 70.69         \\ 
RKD      & 72.28        & 72.21        & 64.43        & 64.43        & 73.21        & 71.50         \\ 
SP       & 73.48        & 74.52        & 68.08        & 68.08        & 74.56        & 73.34         \\ 
PKT      & 74.10        & 72.21        & 66.52        & 66.52        & 74.69        & 73.01         \\ 
CRD      & 75.11        & 76.05        & 69.11        & 69.11        & 75.65        & 74.30         \\
WCoRD    & 75.77        & 76.32        & 70.45        & 69.47        & 75.96        & 74.86         \\ 
SemCKD   & 76.31        & 76.06        & 68.69        & 69.98        & 77.02        & 74.18         \\ 
ReviewKD & 77.45        & 77.14        & 69.89        & 70.37        & 77.78        & 75.34         \\ 
MGD      & 76.22        & 75.89        & 68.54        & 69.44        & 76.65        & 73.89         \\ \toprule
KD       & 74.07        & 74.83        & 67.35        & 67.37        & 74.45        & 73.81         \\ 
SD-KD    & 76.30(\textbf{+2.23}) & 76.65(\textbf{+1.82}) & 69.55(\textbf{+2.10}) & 68.79(\textbf{+1.42}) & 76.67(\textbf{+2.32}) & 74.89(\textbf{+1.08})  \\ 
DKD      & 76.45        & 76.70        & 70.35        & 69.71        & 77.07        & 75.34         \\ 
SD-DKD   & \textbf{77.30(\textbf{+0.85})} & \textbf{77.21(\textbf{+0.51})} & \textbf{71.36(\textbf{+1.01})} & \textbf{70.25(\textbf{+0.54})} & \textbf{78.05(\textbf{+0.98})} & \textbf{75.86 (\textbf{+0.52})} \\

NKD & 75.31 & 75.96 &   69.39  &  68.72 &76.26 &74.01\\ 
SD-NKD & 76.34(\textbf{+0.1.02}) & 76.81(\textbf{+0.85}) & 70.25(\textbf{+0.86}) & 69.50(\textbf{+0.82}) & 77.07(\textbf{+0.81}) & 74.62(\textbf{+0.61})\\

 \toprule
\end{tabular}
\end{center}
\caption{Performance of model compression on the CIFAR-100 dataset. Here, the teacher and student with different network structures but the same layer.}
\label{cifar100_diff}
\end{table*}

\textbf{Training details.} For CIFAR100 and CUB200, our implementation follows the practice in CRD~\cite{CRD}. teachers and students are trained for 240 epochs with SGD. The batch size is 64, the learning rates are 0.01 for ShuffleNet and MobileNet-V2, and 0.05 for the other series (e.g. VGG, ResNet, and WRN). The learning rate is divided by 10 at 150, 180, and 210 epochs. The weight decay and the momentum are set to 5e-4 and 0.9. The weight for distillation loss follows the same weight for KD, DKD, and NKD for a fair comparison.

For ImageNet, we train the models for 100 epochs. As the batch size is 512, the learning rate is initialized to 0.2 and divided by 10 for every 30 epochs. Weight decay is 1e-4. The weight for distillation loss follows the same weight for KD, DKD, and NKD for a fair comparison.

\begin{table*}[h]
\begin{center}
\centering

\begin{tabular}{ccc|cc|cc|cc}
\toprule
   & Teacher & Student & KD  &\textbf{SD-KD} & DKD  & \textbf{SD-DKD} & NKD  & \textbf{SD-NKD}  \\ \toprule[1pt]
Top1 & 73.31  & 69.75  &  70.66 & 71.44(\textbf{+0.84}) & 71.70 & 72.02(\textbf{+0.32}) & 71.96 & \textbf{72.33(\textbf{+0.37})}  \\ 
Top5 & 91.43  & 89.07  & 89.88 & 90.05 & 90.41 & 91.21  & 91.10 & \textbf{91.31} \\ \toprule[1pt]
\end{tabular}
\end{center}
\caption{Top-1 and top-5 accuracy (\%) on the ImageNet validation. We set ResNet-34 as the teacher and ResNet18 as the student.}
\label{imagenet-same}
\end{table*}

\begin{table*}[h]
\begin{center}
\centering

\begin{tabular}{ccc|cc|cc|cc}
\toprule
   & Teacher & Student & KD  &\textbf{SD-KD} & DKD  & \textbf{SD-DKD} & NKD  & \textbf{SD-NKD}  \\ \toprule[1pt]
Top1 & 76.16  & 68.87  &  70.50 & 72.24(\textbf{+1.74}) & 72.05 & 73.08(\textbf{+1.03}) & 72.58 & \textbf{73.12(\textbf{+0.54})}  \\ 
Top5 & 92.86  & 88.76  & 90.34  & 90.71  & 91.05 & 91.09 & 90.80 & \textbf{91.11} \\ \toprule[1pt]
\end{tabular}
\end{center}
\caption{Top-1 and top-5 accuracy (\%) on the ImageNet validation. We set ResNet-50 as the teacher and MobileNet-V1 as the student.}
\label{imagenet-diff}
\end{table*}

\subsection{Comparison Results}

\textbf{Results on the teacher and student with different network structures and layers.} As shown in Table~\ref{diff_all}, SDD consistently contributes to significant performance gains for multiple classical logit distillation methods. Specifically, SDD increases the performance of  ResNet32x4-MobileNetV2, WRN40-2-MobileNetV2, and ResNet50-ShuffleNetV1 by nearly 1\% for KD, DKD, and NKD. These results show the effectiveness of the proposed SDD method in dealing with the teacher and student with different network structures and layers. Moreover, some conventional KD even outperforms the state-of-the-art feature-based methods in some teacher and student pairs, such as WRN-2 and VGG8 as well as ResNet50 and ShufleNetV1. These results show the superiority of logit distillation and the necessity of improving logit knowledge distillation. Here, the result of the ReviewKD method is `-' since it cannot handle such scenarios.


\textbf{Results on the teacher and student with different network structures but the same layer.} As shown in Table~\ref{cifar100_diff}, SDD improves the performance of multiple classical logit distillation methods by 0.5\%$\sim$2.23\% on the CIFAR 100 dataset. Besides, even for the large-scale ImageNet dataset, SDD also brings 0.5\%$\sim$1.74\% performance improvement as shown in Table~\ref{imagenet-diff}. In general, for most teacher-student pairs, SDD can contribute to more than 1\% performance gain on small or large-scale datasets, demonstrating its effectiveness in dealing with the teacher and student with different network structures but the same layer. Besides, the proposed SD-DKD outperforms all the feature-based distillation methods, including the state-of-the-art method ReviewKD and MGD. This further confirms the superiority of SDD.



\textbf{Results on the fine-grained classification task.} As shown in Table~\ref{fine}, SDD improves the performance of multiple classical logit distillation methods by 1.06\%$\sim$6.41\%. These results demonstrate that the proposed SDD achieves more remarkable performance gains on such a task where different classes have small discrepancies. The reason may be that fine-grained classification tasks have a stronger dependence on fine-grained semantic information than conventional classification tasks since different classes have similar global information. At the same time, the SDD can enable the model to capture local information. This demonstrates the potential of the proposed SDD for distilling fine-grained classification models.

\begin{table*}[]
\centering
\begin{tabular}{cccccc}
\toprule
Teacher  & ResNet32x4  & ResNet32x4     & VGG13       & VGG13 & ResNet50     \\ 
Acc      & 66.17       & 66.17          & 70.19       & 70.19 & 60.01        \\ \toprule
Student  & MobileNetV2 & ShuffleNetV1   & MobileNetV2 & VGG8  & ShuffleNetV1 \\ \
Acc      & 40.23       & 37.28          & 40.23       & 46.32 & 37.28        \\ \toprule
SP       &   48.49      &    61.83     & 44.28     &    54.78  &   55.31       \\ 
CRD      &    57.45         &   62.28         & 56.45         &    66.10  &    57.45          \\ 
SemCKD   &    56.89       &    63.78       & 68.23          & 66.54    &   57.20       \\ 
ReviewKD &   -        &    64.12          & 58.66       &  67.10  &      -      \\ 
MGD      &    -      &      -          & -           &  66.89  &       57.12     \\ \toprule
KD       & 56.09       & 61.68          & 53.98       & 64.18 & 57.21        \\ 
SD-KD    & 60.51(\textbf{+4.42}) & 65.46(\textbf{+3.78})  & 59.80(\textbf{+5.82})  & 67.32(\textbf{+3.14}) & 60.56(\textbf{+3.25})        \\
DKD      & 59.94       & 64.51          & 58.45       & 67.20 & 59.21        \\ 
SD-DKD   & \textbf{62.97(\textbf{+3.43})} & \textbf{65.58(\textbf{+1.06})}     & \textbf{64.86(\textbf{+6.41})}  & \textbf{68.67(\textbf{+1.47})} & \textbf{60.66(\textbf{+1.45})}        \\ 
NKD      & 59.81       & 64.0           & 58.40       &    67.16   &    59.11          \\ 
SD-NKD   & 62.69(\textbf{+2.88})       & 65.50(\textbf{+1.5})    & 64.63(\textbf{+6.23})       &    68.37(\textbf{+1.21})   &       60.42(\textbf{+1.31})       \\ \toprule
\end{tabular}
\caption{Performance on the CUB200 dataset. Here, we conduct experiments on three different teacher-student types, the same structure and layer (VGG13-VGG8), different structures while the same layer (ResNet32x4-ShffleNetV1 and VGG13-MobileNetV2), and different structures and layers (ResNet32x4-MobileNetV2 and ResNet50-ShuffleNetV1).}
\label{fine}
\end{table*}

\subsection{Ablation Study}

The ablation studies are conducted on CIFAR-100 by using the heterogeneous teacher-student pair ResNet32x4/ShuffleNetV1 and homogeneous pair ResNet32x4/ResNet8x4, respectively.

\textbf{Effect of the decoupled logit knowledge}. As shown in Table~\ref{wrong-ringht-local}, the fusion of consistent and complementary logit knowledge improves the performance of conventional KD by 3.30\% and 2.23\% for homogeneous and heterogeneous teacher-student pairs, respectively. This demonstrates the effectiveness of decoupling the global logit knowledge into multiple semantic-independent logit knowledge. Besides both consistent and complementary logit knowledge consistently improve the performance of conventional KD for homogeneous and heterogeneous teacher-student pairs, respectively, verifying their effectiveness.

\begin{table}[]
\centering
\begin{tabular}{c|c|c}
\toprule
         & \begin{tabular}[c]{@{}c@{}} ResNet32x4-\\ ResNet8x4\end{tabular} & \begin{tabular}[c]{@{}c@{}} ResNet32x4-\\ShuffleNetV1\end{tabular} \\ \toprule
 N/A     & 73.33                & 74.07 \\
Consistent    & 75.14                & 75.88                   \\ 
Complementary & 74.79                & 75.10                   \\ 
Fusion   & \textbf{76.63}       & \textbf{76.30}          \\ \toprule
\end{tabular}
\caption{Performance of the SDD method that uses different decoupled logit knowledge on the teacher-student pairs with homogeneous and heterogeneous network structures. `N/A' denotes the result of the conventional KD.}
\label{wrong-ringht-local}
\end{table}

\textbf{Effect of different decoupled scales.} We set different scale sets ${M}$ for the $L_{SDD} $ to study the effect of local logit knowledge at different scales. Specifically, each scale $m$ in ${M}$ means the logit knowledge of region $\mathcal{Z}(m,n)$ is utilized. For a fair comparison, we take the model only using the global knowledge as the baseline, namely $M=\{1\}$. The results are shown in Table~\ref{multi_scale}.


For the teacher-student pair of ResNet32x4 and ShuffleNetV2, SDD achieves the best results when $M=\{1,2,4\}$.~In contrast, the teacher-student pair of ResNet32x4 and ResNet8x4 reaches the best results when ${M}=\left\{1,2\right\}$. This indicates that the teacher-student pair with heterogeneous structures needs more fine-grained semantic knowledge than the teacher-student pair with homogeneous structures. This may be because the structure heterogeneity is not conducive for students to imitate the teacher's knowledge. Besides, because the loss $L_{SDD}$ accumulates for all scales, too many scales for homogeneous teacher-student pairs will bring redundant information since their feature maps are similar. This prevents students from capturing key scales and degrades their performance. Therefore, scale set ${M}$ of SD-KD, SD-DKD, and SD-NKD is set as $\left\{1,2,4\right\}$ for the distillation tasks with structure discrepancy, and $\left\{1,2\right\}$ for the distillation tasks with similar structures.





\begin{table}[h]
\centering

\begin{center}
  
\begin{tabular}{c|c|c}
\toprule[1pt]
            & \begin{tabular}[c]{@{}c@{}}ResNet32x4-\\ResNet8x4\end{tabular} & \begin{tabular}[c]{@{}c@{}}ResNet32x4-\\ShuffleNetV1\\\end{tabular} \\ \toprule[1pt]
M=\{1\}     &73.33                                                            &      74.07                                                      \\ 
M=\{1,2\}   & \textbf{76.63}                                                             &     75.10                                                      \\ 
M=\{1,4\}   &   76.13                                                           &        75.84                                                   \\ 
M=\{1,2,4\} & 75.74                                                              &                        \textbf{76.30}                                   \\ \toprule[1pt]
\end{tabular}
\end{center}
\caption{Performance of the SDD method that uses local logit knowledge at different scales on the teacher-student pairs with homogeneous and heterogeneous network structures. The $\mathcal{LD}$ used for $L_{SDD}$ is conventional logit KD loss.}
\label{multi_scale}
\end{table}

\textbf{Effect of $\beta$}. The results are shown in Table~\ref{para}. Firstly, all $\beta$ greater than 1 obtain the performance gain compared to $\beta=1$. This demonstrates the effectiveness of paying more attention to ambiguous samples. Besides, the model achieves the best performance when $\beta=2$. Thus, we set $\beta$ as $2.0$ in the experiments.

\begin{table}[]
\centering
\begin{tabular}{c|ccccc}

$\beta$   & 1     & 2              & 4     & 6              & 8     \\ \hline
Acc & 75.62 & \textbf{76.30} & 75.99 & 76.15          & 75.78 \\ 
\end{tabular}
\caption{Performance of SDD method with different $\beta$ on  ResNet32×4 and ShuffleNetV1.}
\label{para}
\end{table}



\textbf{Training efficiency.} We assess the training time of state-of-the-art distillation methods to evaluate the efficiency of SDD. As shown in Table~\ref{EC}, the training time of SD-KD is the same as the KD and less than the feature-based methods. This is because SDD calculates the multi-scale logit output via the same classifier, introducing no extra structural complexity. These results demonstrate the computational efficiency of SDD.

\begin{table}[ht]
    \centering
    \begin{tabular}{ccccc}
    \toprule
        Methods  & CRD & ReviewKD & KD & SD-KD \\ \toprule
        Times(ms) & 41 & 26 & 11 & 11 \\ \bottomrule
    \end{tabular}
    \caption{Training time (per batch) vs. accuracy on CIFAR-100. We set ResNet32×4 as the teacher and ResNet8×4 as the student.}
    \label{EC}
\end{table}


  


\begin{figure}[t]
\centering
\includegraphics[width=1.0\columnwidth]{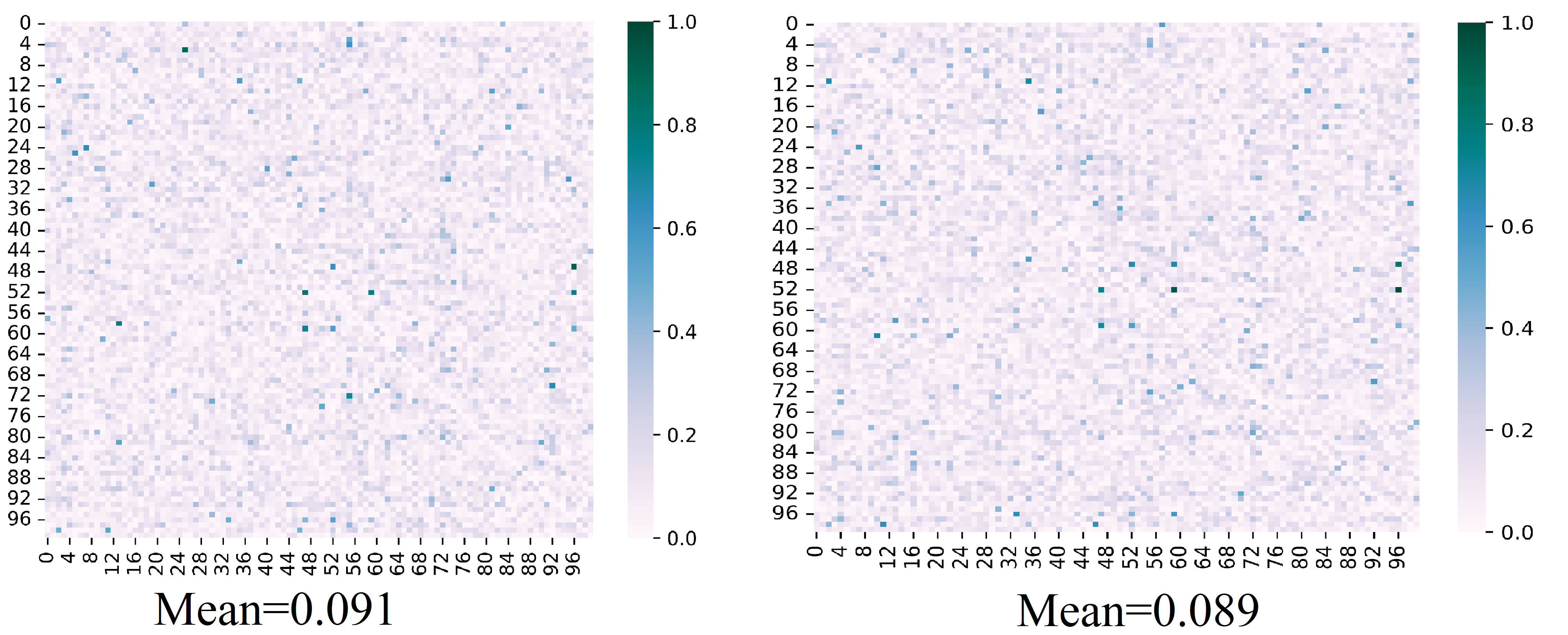} 
\caption{Difference of correlation matrices of student and teacher logits of KD (left) and SD-KD (right).}
\label{corrent}
\end{figure}
\begin{figure}[t]
\centering
\includegraphics[width=1.0\columnwidth]{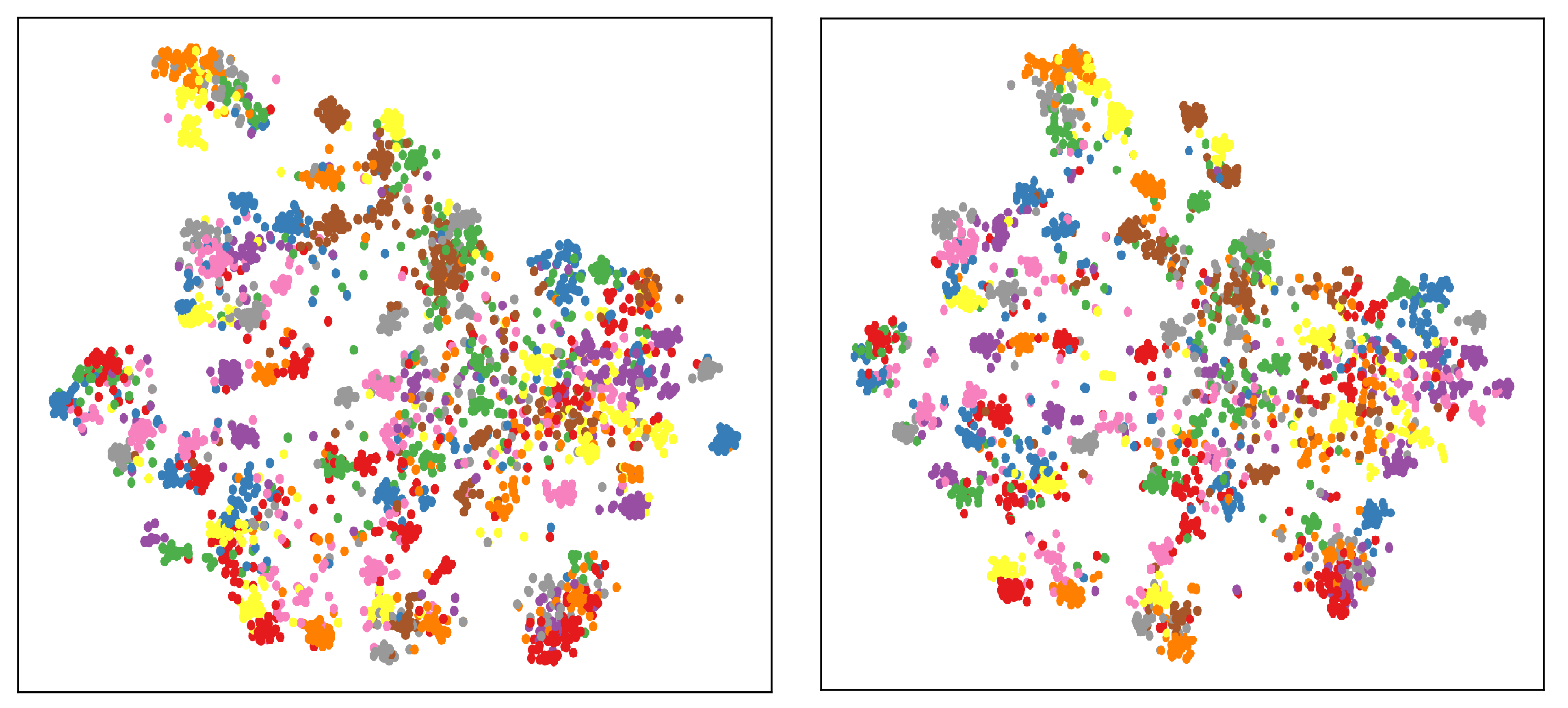} 
\caption{t-SNE of features learned by KD (left) and SD-KD (right).}
\label{feature_projection}
\end{figure}
\begin{figure}[t]
\centering
\includegraphics[width=1.0\columnwidth]{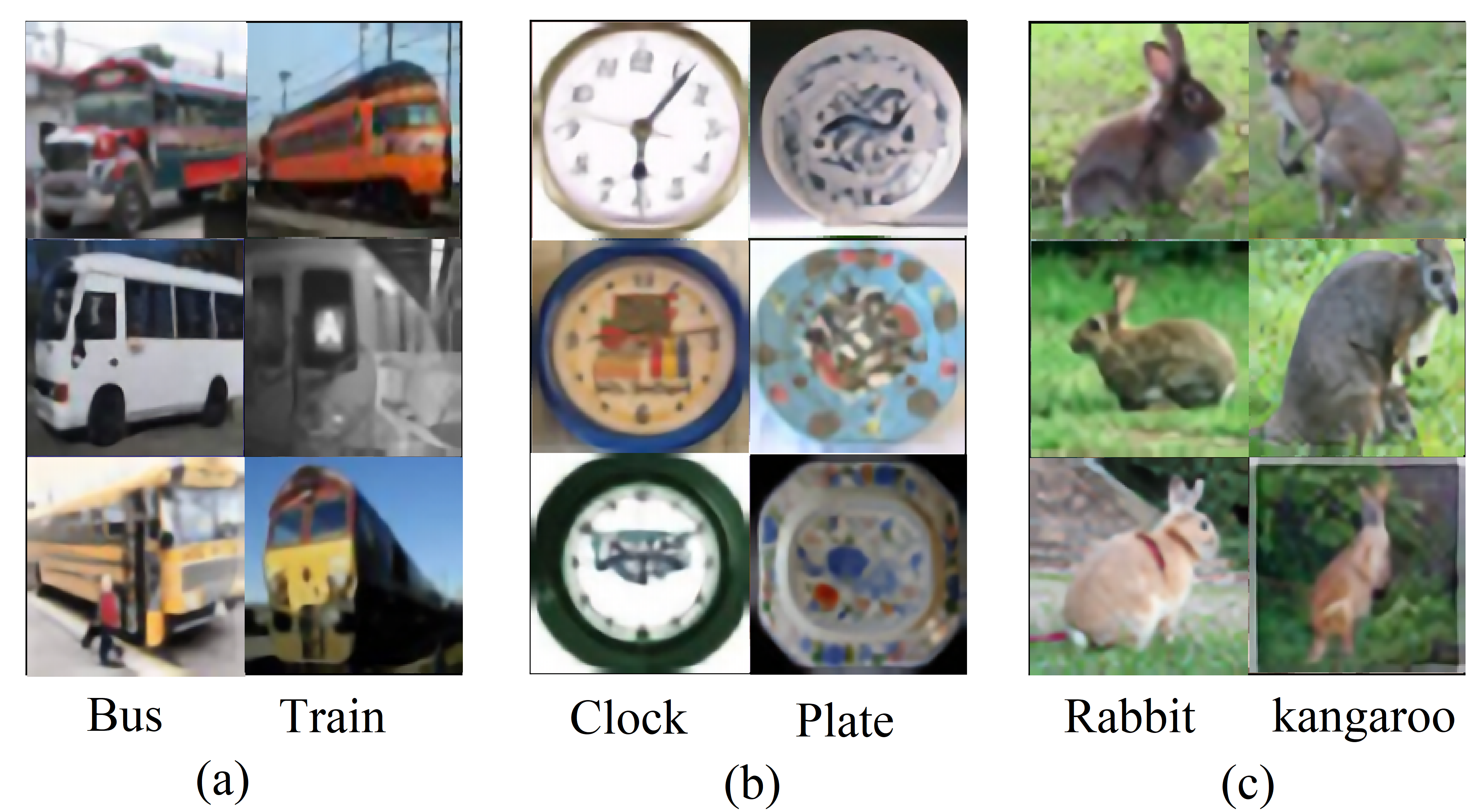} 
\caption{Some examples that can be classified correctly by the student trained with SD-KD while misclassified by the student trained with conventional KD.}
\label{sample}
\end{figure}

\textbf{Visualizations.} We present visualizations from three perspectives (with setting teacher as ResNet32x4 and the student as ResNet8x4 on CIFAR-100). (1) Fig.~\ref{corrent} shows the difference of correlation matrices of the global logits of student and teacher. Different from the DKD method, SD-KD provides a similar difference map as the KD. This indicates that the improvement of SDD does not come from better imitating the global logit output of the teacher. (2) To study the mechanism of SDD, we visualize the inter-class distance map via t-SNE projection in Fig.~\ref{feature_projection}. We can see that the representations of SD-KD are more separable than KD, showing the proposed SSD can enhance the discriminative ability of students. (3) In addition, we further visualize some cases that can be classified correctly by the student trained with SD-KD while misclassified by the student trained with conventional KD (Fig.~\ref{sample}).  From the figure, we can see that the samples misclassified by the KD model are exactly the ambiguous samples that seem similar in the global semantics. These results verify our proposition that SDD can help the student acquire the fine-grained semantic information of local regions to regularize global knowledge, improving its discrimination ability to the ambiguous samples.

\section{Conclusion}
\label{cl}

This paper revisits conventional logit-based distillation and reveals that the coupled semantic knowledge in global logit knowledge limits its effectiveness. To overcome this limitation, we propose Scale-Decoupled Knowledge Distillation (SDD), which decouples the global logit output into multiple local logit outputs. We establish knowledge-transferring pipelines for these outputs to transfer fine-grained and unambiguous semantic knowledge. In addition, we can further divide the decoupled knowledge into consistent and complementary parts that transfer the semantic information and sample ambiguity, respectively. SDD can guide the student to pay more attention to ambiguous samples by increasing the weight of complementary parts, improving its discrimination ability for ambiguous samples. Extensive experiments on several benchmark datasets demonstrate the effectiveness of SDD across a range of teacher-student pairs, particularly in fine-grained classification tasks.

\section{Appendix}


\subsection{Ablation Study}

\textbf{Combination of the proposed logit distillation and other feature distillation}

\begin{table}[h]
\setlength\tabcolsep{5pt} 
\begin{tabular}{cccc}
\hline
Teacger   & WRN40-2 & ResNet32x4  & ResNet32x4 \\
Student   & VGG8    & ShufleNetV1 & ResNet8x4  \\ 
SP    &73.18 & 73.48 &72.94 \\
CRD     & 73.88 & 75.11 & 75.51 \\
SDD-KD     &    74.44     &     76.30        &      75.63      \\ 
SDD-KD+SP  &    74.02     &      75.87       &       76.33     \\ 
SDD-KD+CRD &    \textbf{74.82}     &      \textbf{76.89}       &    \textbf{76.94}        \\ \hline
\end{tabular}
\end{table}

We consider the teacher and student to be heterogeneous in layer and structure (column 1), heterogeneous in structure (column 2), and homogeneous (column 3). In general, the combination of the proposed logit distillation and feature distillation can improve the model performance (see MS-KD+CRD), but not all of them can bring improvement (see MS-KD+SP). This may be because SP is a more early method and has a huge performance gap with MS-KD.

\textbf{The performance of directly aligning the features after multi-scale pooling without using the linear classifier} The experiment results are shown below.

 \begin{table}[h]
 \centering
 \vspace{-0.6em}
\begin{tabular}{c|c|c}
\hline
      & \begin{tabular}[c]{@{}c@{}}ResNet32x4\\ -ResNet8x4\end{tabular} & \begin{tabular}[c]{@{}c@{}}ResNet32x4-\\ ShuffleNetV2\end{tabular} \\ \hline
SP    & 72.94                                                           & 74.45                                                              \\ 
SDD-SP & 74.37                                                           & 75.65                                                              \\ 
SDD-DKD & \textbf{76.63}                                                           & \textbf{78.05 }                                                          \\ \hline
\end{tabular}
\vspace{-1.1em}
\end{table}

SD-SP means performing SP distillation on the multi-scale pooling feature directly without using the linear classifier. These results show that the scale decoupling operator is effective for logit-based distillation, and also for feature-based, while the former has higher performance gain.

\textbf{Prediction and image visualization on CIFAR 100}

\begin{figure}[t]
\centering
\includegraphics[width=1.0\columnwidth]{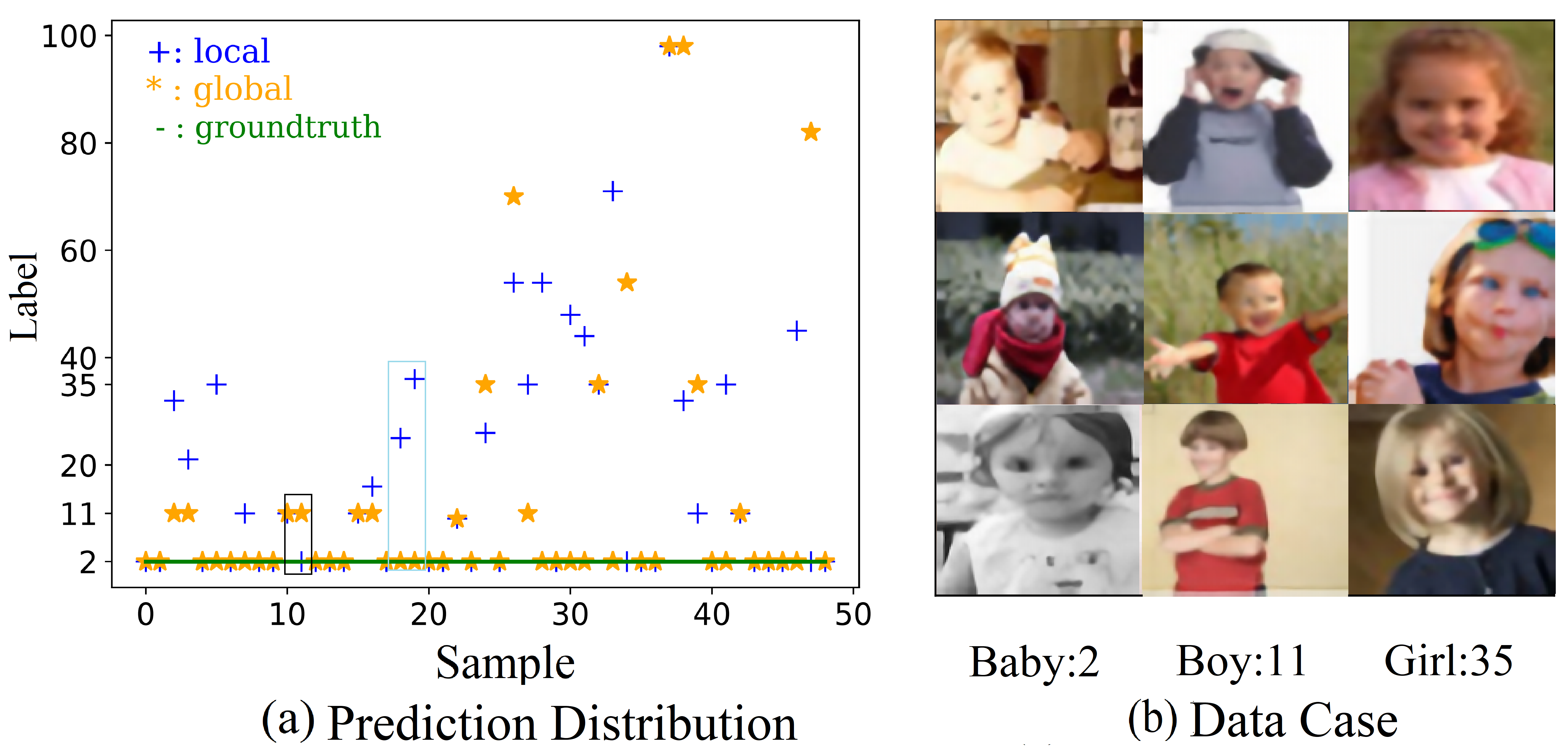} 
\caption{Prediction and image visualization on CIFAR 100. (a) shows the multi-scale prediction results of ResNet32X4 from the baby samples, which are mainly distributed among `2' (groundtruth), `11'(boy), and `35'(girl). (b) shows some examples of baby, boy, and girl images.}
\label{case_prediction_cifar100}
\end{figure}

As shown in Fig.~\ref{case_prediction_cifar100}, the misclassification usually occurs in the samples with similar global semantics. For example, the baby itself also contains the semantics of little boy and girl. And the multi-scale prediction may be inconsistent in the ambiguous samples and the local logit knowledge can preserve fine-grained semantic information to regularize the global logit knowledge, assisting the training of the student network. On the one hand, when the global prediction is right while the local prediction is wrong (see the cyan box in Fig.~\ref{case_prediction_cifar100}(a)), the inconsistent local knowledge encourages the student to preserve the sample ambiguity, avoiding overfitting the ambiguous samples. On the other hand, when the global prediction is wrong while the local prediction is right (see the block box in Fig.~\ref{case_prediction_cifar100}(a)), the inconsistent local knowledge can encourage the student to learn from similar components among different categories, alleviating the bias caused by the teacher.

\subsection{Discussion}
\textbf{The difference between feature-based distillation and SDD}
While SDD requires the information of the convolution layer, it is still the logit-based distillation method since it transfers knowledge via logit output. It has the advantages belonging to the logit-based distillation, such as computational efficiency (see Table 9).

{
    \small
    \bibliographystyle{ieeenat_fullname}
    \bibliography{main}
}


\end{document}